%% file: main.tex
\documentclass[fleqn,10pt]{wlscirep}
\usepackage[utf8]{inputenc}
\usepackage[T1]{fontenc}
\usepackage{amsmath}

\title{DLBCL-Morph: Morphological features computed using deep learning for an annotated digital DLBCL image set}
\author[1,$\dag$]{Damir Vrabac}
\author[1,$\dag$]{Akshay Smit}
\author[2]{Rebecca Rojansky}
\author[2]{Yasodha Natkunam}
\author[3]{Ranjana H. Advani}
\author[1]{Andrew Y. Ng}
\author[2,$\ddag$]{Sebastian Fernandez-Pol}
\author[1,$\ddag$,*]{Pranav Rajpurkar}
\affil[1]{Department of Computer Science, Stanford University}
\affil[2]{Department of Pathology, Stanford University School of Medicine}
\affil[3]{Department of Medicine, Division of Oncology, Stanford University School of Medicine}
\affil[*]{Corresponding author(s): Pranav Rajpurkar (pranavsr@stanford.edu)}
\affil[$\dag$]{These authors contributed equally to this work.}
\affil[$\ddag$]{These authors contributed equally to this work.}

\begin{abstract}
Diffuse Large B-Cell Lymphoma (DLBCL) is the most common non-Hodgkin lymphoma. Though histologically DLBCL shows varying morphologies, no morphologic features have been consistently demonstrated to correlate with prognosis. We present a morphologic analysis of histology sections from 209 DLBCL cases with associated clinical and cytogenetic data. Duplicate tissue core sections were arranged in tissue microarrays (TMAs), and replicate sections were stained with H\&E and immunohistochemical stains for CD10, BCL6, MUM1, BCL2, and MYC. The TMAs are accompanied by pathologist-annotated regions-of-interest (ROIs) that identify areas of tissue representative of DLBCL. We used a deep learning model to segment all tumor nuclei in the ROIs, and computed several geometric features for each segmented nucleus. We fit a Cox proportional hazards model to demonstrate the utility of these geometric features in predicting survival outcome, and found that it achieved a C-index (95\% CI) of 0.635 (0.574,0.691). Our finding suggests that geometric features computed from tumor nuclei are of prognostic importance, and should be validated in prospective studies.

\end{abstract}
\begin{document}

\flushbottom
\maketitle

\thispagestyle{empty}

\begin{figure}[ht]
\centering
\includegraphics[width=\linewidth]{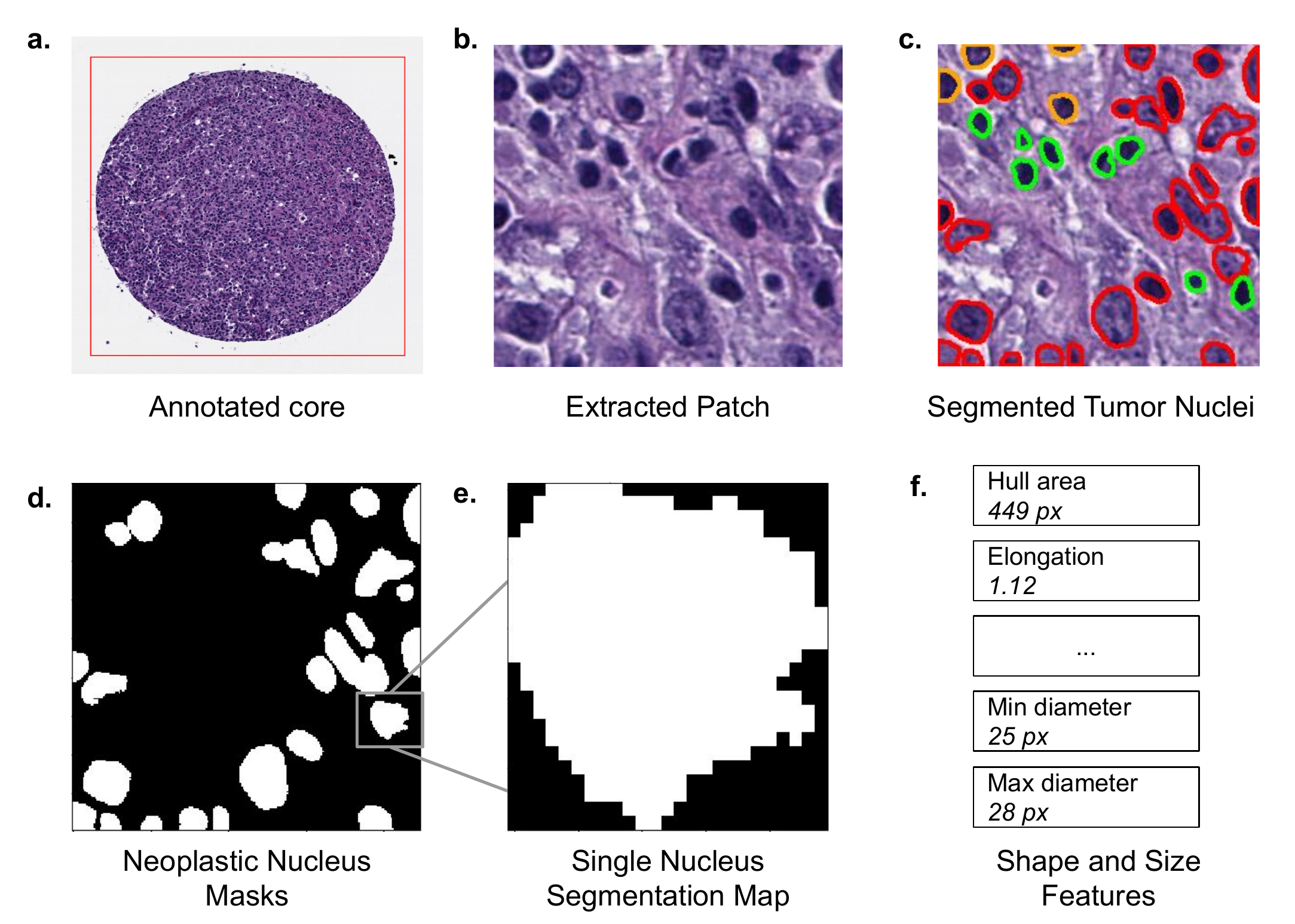}
\caption{\textbf{Data pipeline for a single core from an H\&E stained tissue microarray (TMA).} In a) the red rectangle is the pathologist-annotated ROI. In c) red corresponds to cell nuclei classified as ``neoplastic" by HoVer-Net. Green corresponds to ``inflammatory" and orange corresponds to ``non-neoplastic epithelial".}
\label{fig:workflow}
\end{figure}

\input{Sections/background}
\input{Sections/methods}

\input{Sections/data_records}
\input{Sections/technical_validation}
\input{Sections/usage_notes}
\input{Sections/code_availability}

\bibliography{bibliography}




\section*{Author contributions statement}

DV, AS, SF, and PR developed the concept and design;  DV, AS, RR, YN, RHA, SF, and PR performed acquisition, analysis, or
interpretation of data; AYN, SF, and PR provided supervision. DV, AS, and RR drafted the manuscript, and all authors
provided critical revision of manuscript for important intellectual content.

\section*{Competing interests}

The authors declare no competing interests.


\end{document}

%% file: Sections/background.tex
\section*{Background \& Summary}

Diffuse Large B-Cell Lymphoma (DLBCL) is the most common type of non-Hodgkin lymphoma (NHL), accounting for over a third of cases \cite{1997} with more than 20,000 patients diagnosed annually in the United States \cite{Horvat2018}. DLBCL is fatal without treatment, however approximately 70\% of patients can be cured with contemporary therapeutic regimens \cite{Leonard2017}. Treatment outcomes following standard R-CHOP (rituximab, cyclophosphamide, doxorubicin, vincristine, and prednisone) therapy are highly variable, and depend on a number of clinical, biologic, and genetic factors. Currently, the most effective prognostic classification is the National Comprehensive Cancer Network International Prognostic Index (NCCN-IPI), which incorporates five clinical variables including age, lactate dehydrogenase (LDH), extra-nodal sites of involvement, Ann Arbor stage, and ECOG performance status \cite{Zhou2014}. The NCCN-IPI model is widely used to risk stratify patients into good, intermediate, and poor-risk categories, however it is insufficient to guide therapeutic decision-making for individual patients.
 
Gene expression profiling (GEP) studies revealed distinct subtypes of DLBCL that correspond to differences in cell of origin (COO) and show different outcomes in response to R-CHOP therapy \cite{Alizadeh2000, Scott2015}. This approach categorizes DLBCL as either germinal center B-cell (GCB), activated B-cell (ABC), or indeterminate, based on the phase of B-cell development it most closely resembles \cite{Basso2015}. A practical algorithm employing this approach for immunohistochemically stained, formalin fixed, paraffin embedded tissue was developed by Hans et al, and despite imperfect concordance with the gold standard GEP method, it is now the most widely used algorithm in the United States for DLBCL \cite{Riedell2018}. The GCB subtype is associated with more favorable outcomes than the non-GCB subtype \cite{Alizadeh2000, GutirrezGarca2011, Scott2015_2, Fu2008, Alizadeh2011, Lenz2008}. 
 
In addition to COO subtyping, double-hit lymphomas with concurrent chromosomal translocations of the MYC and BCL2 genes, or less commonly MYC and BCL6 genes, and double-expressor lymphomas with dual overexpression of MYC and BCL2 proteins have been found to correlate with an aggressive clinical course and poor outcomes when treated with R-CHOP \cite{Riedell2018_2}. Determination of these molecular subsets is now standard of care per the World Health Organization (WHO) guidelines and patients harboring dual chromosomal translocations are now formally classified as having high grade B-cell lymphoma, with MYC and BCL2 and/or BCL6 translocations (HGBL) \cite{10.1182/blood-2016-01-643569}.  
 
While COO subtyping by the Hans algorithm corresponds to morphologically distinct benign precursors, germinal center type B-cells and activated B-cells, classification based on the morphologic properties of the tumor itself has historically been challenging due to the significant histomorphologic heterogeneity of DLBCL. Cytologically, DLBCL may resemble centroblasts with multiple peripheral nucleoli and vesicular chromatin or immunoblasts with abundant cytoplasm and a single prominent nucleolus. However, the prognostic significance of these and other recognised cytologic variants, for example anaplastic type DLBCL, is unclear and the subject of continued debate \cite{10.1182/blood.V89.7.2291, Baars1999, Diebold2002, Nakamine1993, Salar2009, Villela2001}.

Though several studies have thus far failed to conclusively demonstrate that morphologic classification can predict outcomes in DLBCL, automated imaging methods could potentially identify novel, prognostically significant morphological or immunohistochemical biomarkers. The ability of automated methods to identify prognostically relevant features on H\&E sections that have eluded pathologists has been demonstrated \cite{Beck108ra113, Kather2019, Jain2020.06.15.153379}. If successful, automated image analysis could be scaled up into a cost-effective alternative to current classification methods which are typically costly and/or labor intensive. A critical requirement for the development of these models is the availability of datasets containing digitally scanned slides stained to show cell morphology and expression of relevant proteins with accompanying prognostic outcome data.

Here we present DLBCL-Morph, a publicly available dataset containing 42 digitally scanned high-resolution tissue microarrays (TMAs) from 209 DLBCL cases at Stanford Hospital. Each TMA was stained for H\&E as well as for  CD10, BCL6, MUM1, BCL2, and MYC protein expression. All of the TMAs are accompanied by pathologist-annotated regions-of-interest (ROIs) that indicate areas representative of DLBCL. For each patient in the dataset, we provide survival data, follow-up status, and a wide range of clinical and molecular variables such as age and MYC/BCL2/BCL6 gene translocations. We also segmented out tumor nuclei from ROIs inside the H\&E stained TMAs, and provide several geometric features for each tumor nucleus. 

%% file: Sections/methods.tex
\begin{figure}[ht]
\centering
\includegraphics[width=\linewidth]{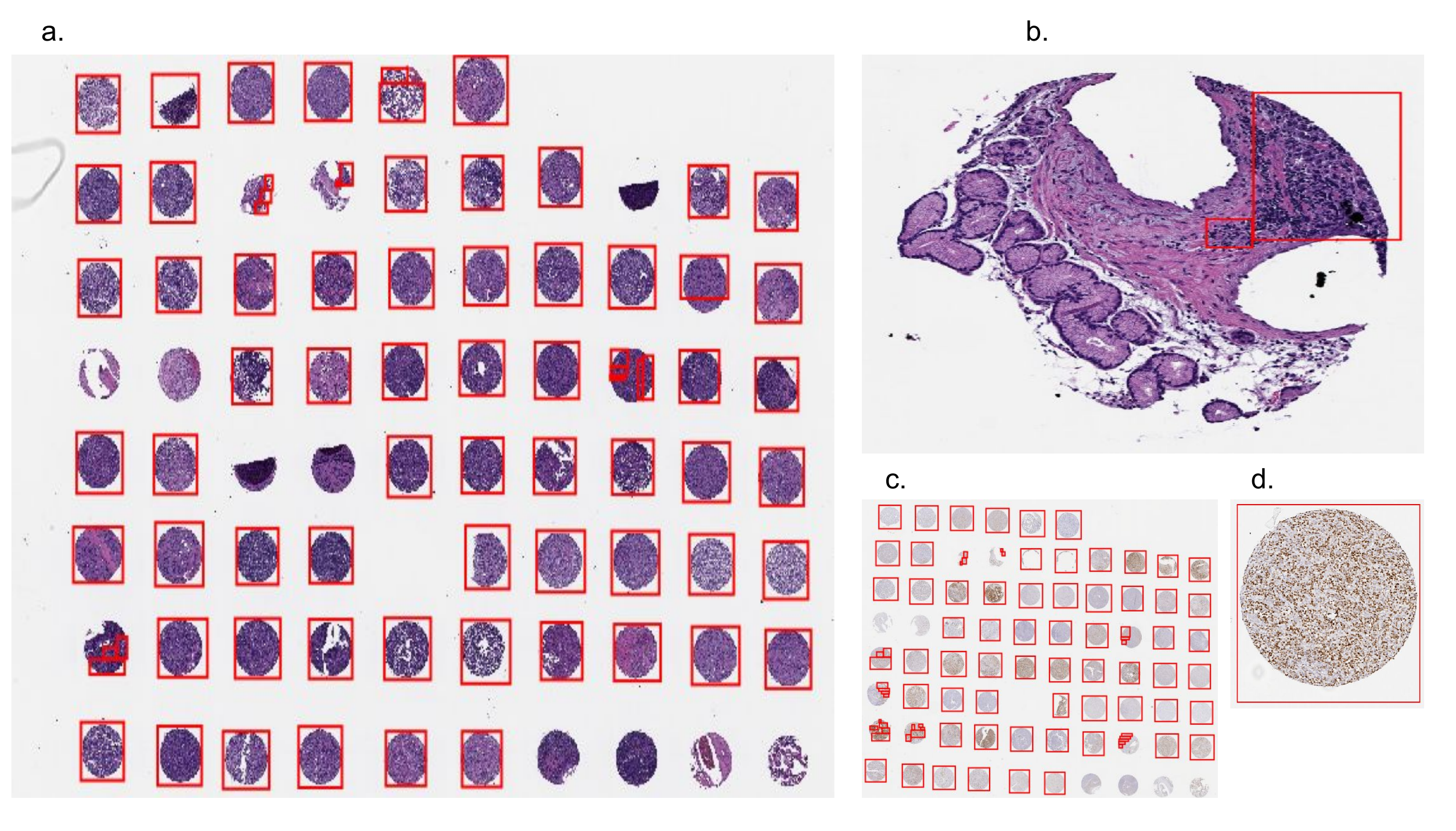}
\caption{\textbf{Tissue microarrays (TMAs) with region-of-interest (ROI) annotations.} a) H\&E stained TMA. The red rectangles denote ROIs annotated by a human expert. Some missing or unrepresentative cores have no ROIs. b) A single core from the TMA in a) with ROI that ignores unrepresentative areas of the core. c) BCL6 stained TMA, containing cores from the same patients as a). d) A single annotated core from the TMA in c). Cells stained orange show greater BCL6 expression.}
\label{fig:he_roi}
\end{figure}

\section*{Methods}

Our dataset contains digitally scanned TMAs accompanied by pathologist-annotated ROIs. We extracted patches from the ROIs inside the H\&E stained TMAs, and used a deep learning model called HoVer-Net \cite{graham2018hovernet} to segment tumor cell nuclei. We then computed several geometric descriptors for each segmented nucleus. Figure \ref{fig:workflow} shows our pipeline for an H\&E stained TMA core. Our project was approved by the Institutional Review Board of Stanford University. All protected health information was removed and the project had no impact on clinical care, so the requirement for individual patient consent was waived. 

\subsection*{Patient Cohort}

The study cohort consists of patients with de novo, CD20+ DLBCL treated with curative intent with R-CHOP or R-CHOP–like immunochemotherapy with available clinical data from the Stanford Cancer Institute, Stanford, California. This patient cohort was included in a prior study with clinicopathologic inclusion criteria are as previously described \cite{doi:10.1200/JCO.19.00743}.

\subsection*{Tissue Microarray} Stained tissue microarray (TMA) slides were scanned at 40x magnification (0.25 $\mu$m per pixel) on an Aperio AT2 scanner (Leica Biosystems, Nussloch, Germany) in ScanScope Virtual Slide (SVS) format. This high magnification level displays the tissue in very fine detail, which we believe to be beneficial for the development of automated imaging models. Each SVS file also contains a slide label image, a macro camera image, and a thumbnail image. The slide label image is a low-resolution image of the slide’s label, which shows the TMA number and the stain (eg: BCL2). The macro camera image is a low-resolution picture of the entire slide. The thumbnail is an image of the whole scanned TMA.

Our dataset includes 7 TMAs, each with a 0.4 micron thick formalin-fixed, paraffin-embedded (FFPE) section of tumors assembled in a grid. Within the microarray each tumor is represented by a 0.6-mm core diameter sample in duplicate. Replicates of each TMA were stained with H\&E, which shows cell morphology. They were also stained for the expression of the following 5 oncogenes: CD10, BCL6, MUM1, BCL2, and MYC. We therefore have 6 stains per TMA, resulting in 42 distinct digitally-scanned slides. An example of an H\&E stained TMA is shown in Figure \ref{fig:he_roi} a) and a BCL6 stained TMA is shown in Figure \ref{fig:he_roi} c).  Since overexpression of one or more of these proteins is observed in a significant portion of DLBCL cases, automated imaging models can use the immunostained TMAs to potentially identify prognostically significant features related to protein expression.

\begin{figure}[t]
\centering
\includegraphics[width=\linewidth]{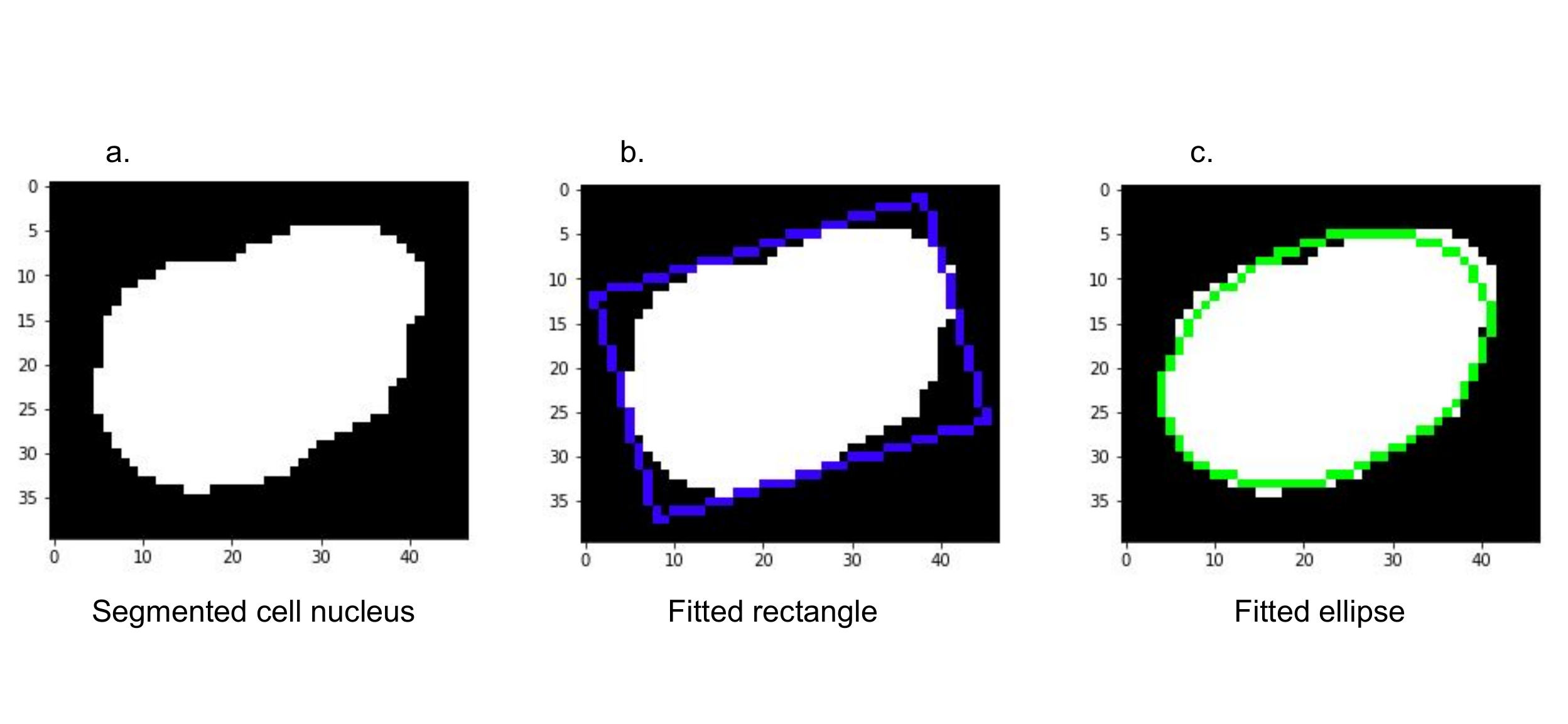}
\caption{\textbf{Rectangle and ellipse fitted to a single segmented tumor nucleus.} a) a binary segmentation image for a tumor cell nucleus. For visual clarity, the image is zero-padded by 5 pixels on each side. b) rotated rectangle fit to the nucleus. Our dataset provides the rectangle's center coordinates, width, height and rotation angle. c) rotated ellipse fit to the nucleus. Our dataset provides the ellipse's center coordinates, perimeter, area, and major and minor axis lengths.}
\label{fig:rect_ellip}
\end{figure}

\subsubsection*{Pathologist annotations}
 
Although TMA cores were already taken from areas of tissue showing DLBCL, some of the cores were partially or entirely missing. Furthermore, some cores still contained areas of tissue that had very few or no tumor cells. We obtained rectangular ROI annotations from expert pathologists to highlight the core regions which represent DLBCL accurately. The annotations were created for all TMAs and all stains at 40x magnification. The pixel coordinates for the rectangles in ROIs, along with the corresponding deidentified unique patient\_id, are included in our dataset. We believe the exclusion of missing or insufficiently representative tissue areas will be beneficial for automated prognostic models which use patches from the TMAs as input. Example ROI annotations are shown in Figure \ref{fig:he_roi} b) and d).

\subsubsection*{Patches from stained TMAs} 

We extracted patches of size 224x224 from within the ROIs in the stained TMAs, at 40x magnification. The patches were extracted uniformly from inside each annotated rectangle, starting from the top-left corner and proceeding until the bottom-right corner. The patches are non-overlapping, and we omitted patches that are mostly white and contain little tissue. We provide these patches as part of our dataset. Due to our ROI annotation process detailed above, our patches exclude missing and unrepresentative areas of cores. Since deep learning based imaging methods typically cannot directly operate on images as large as the 40x magnification image, the patches can instead be used as input. We also used patches from H\&E stained TMAs to segment tumor cell nuclei as described below. 

\subsubsection*{Tumor cell nucleus segmentation} We used a deep learning based nucleus segmentation and classification model called HoVer-Net to segment every tumor cell inside each of the patches from H\&E stained TMAs. The HoVer-Net operates independently on each patch, and produces an output image segmenting all individual cell nuclei in the patch, and another output image specifying the classification of each segmented nucleus. The HoVer-Net classifies segmented nuclei into 5 categories: neoplastic, non-neoplastic epithelial, inflammatory, connective, dead. HoVer-Net uses a neural network based on a pretrained ResNet-50 architecture to extract image features. These extracted features are then processed in three steps: the nuclear pixel (NP) step, HoVer step, and nuclear classification (NC) step. The NP step determines whether each pixel belongs to a nucleus or the background, and the HoVer step predicts the vertical and horizontal distances of nucleus pixels to their centroid, thereby allowing separation of touching nuclei. Then the NC step classifies each nucleus pixel, and aggregates these across all pixels in a segmented nucleus to classify each nucleus as neoplastic, non-neoplastic epithelial, inflammatory, connective, or dead.  We used the HoVer-Net output to identify each neoplastic cell nucleus in a patch, and saved it as a separate binary image, thereby obtaining one binary image for each tumor cell. Each binary image illustrates the size and shape of the nucleus, and we provide these in our dataset. An example binary image is shown in Figure \ref{fig:workflow} e) and another is shown in Figure \ref{fig:rect_ellip} a). We used these binary images to compute geometric features for each tumor cell nucleus as described below. 

\subsubsection*{Geometric features from tumor nuclei} We used the per-nucleus binary segmentation images to compute several geometric features for each tumor cell nucleus. While end-to-end imaging models may not require such hand-crafted features, prognostic models which use these features can give more explainable results, and can more clearly indicate the prognostic importance of these features. 

We fit a (possibly rotated) rectangle of minimum area enclosing the binary mask, and provide the rectangle's top left point coordinates, width and height, and rotation angle. An example rectangle is shown in Figure \ref{fig:rect_ellip} b). The rectangle's top left point is a tuple corresponding to the feature rectCenter. The first element of the tuple corresponds to the x-coordinate, and the second element corresponds to the y-coordinate. The width and height are in a tuple corresponding to the feature rectDimension. The first element of the tuple corresponds to the width, and the second element to the height. The rotation angle corresponds to the feature rotate\_angle, which ranges from $-90^\circ$ to $0^\circ$. A value of $-90^\circ$ corresponds to an axis-aligned rectangle. As the rectangle is rotated clockwise, the angle increases toward $0^\circ$, at which point the rectangle is again axis-aligned and the angle resets to $-90^\circ$.

We fit an ellipse around the nucleus in the binary segmentation mask, and provide the ellipse center, major axis, minor axis, perimeter and area of the ellipse. An example ellipse is shown in Figure \ref{fig:rect_ellip} c). The ellip\_centroid parameter is a tuple containing the x and y coordinates of the ellipse. The features shortAxis and longAxis correspond to the lengths of the minor and major axes respectively. The feature ellip\_perimt corresponds to the ellipse perimeter, and ellip\_area corresponds to the ellipse area. 

We computed the maximum and minimum Feret diameters for each segmented nucleus, and provide the corresponding angles. Given an object and a fixed direction, the Feret diameter is the distance between two parallel tangents to the object, where the tangents are perpendicular to the fixed direction. The feature maxDiameter contains the Feret diameter maximized over all directions, and maxAngle specifies the angle (between $-180^\circ$ and $180^\circ$) at which the maximum diameter is obtained. The features minDiameter and minAngle are similar but for the minimum Feret diameter. We further computed the convex hull of the segmented nucleus. The feature hull\_area corresponds to the area of the convex hull. 

Finally we computed a number of geometric features derived from the quantities described above. These features are esf, csf, sf1, sf2, elongation, and convexity. These are defined below in $(1) - (6)$. The esf, sf1, sf2 and elongation are all simple ratios that can be thought of as measures of how ``elongated" the nucleus is. In particular, they are all equal to $1$ if the nucleus is perfectly circular. The csf is similar: it is a measure of circularity, and is equal to $1$ if the nucleus is perfectly circular. For increasingly elliptical nuclei, the csf decreases towards $0$. 

\begin{equation} 
\text{esf} = \frac{\text{shortAxis}}{\text{longAxis}}
\end{equation}
\begin{equation} 
\text{csf} = \frac{4 \pi * \text{ellip\_area}}{\text{ellip\_perimt}^2}
\end{equation}
\begin{equation}
\text{sf1} = \frac{\text{shortAxis}}{\text{maxDiameter}}
\end{equation}
\begin{equation}
\text{sf2} = \frac{\text{minDiameter}}{\text{maxDiameter}}
\end{equation}
\begin{equation}
\text{elongation} = \frac{\text{maxDiameter}}{\text{minDiameter}}
\end{equation}
\begin{equation}
\text{convexity} = \sqrt{\frac{\text{ellip\_area}}{\text{hull\_area}}}
\end{equation}

%% file: Sections/data_records.tex
\section*{Data Records}

\begin{figure}[h]
\centering
\includegraphics[width=0.4\linewidth]{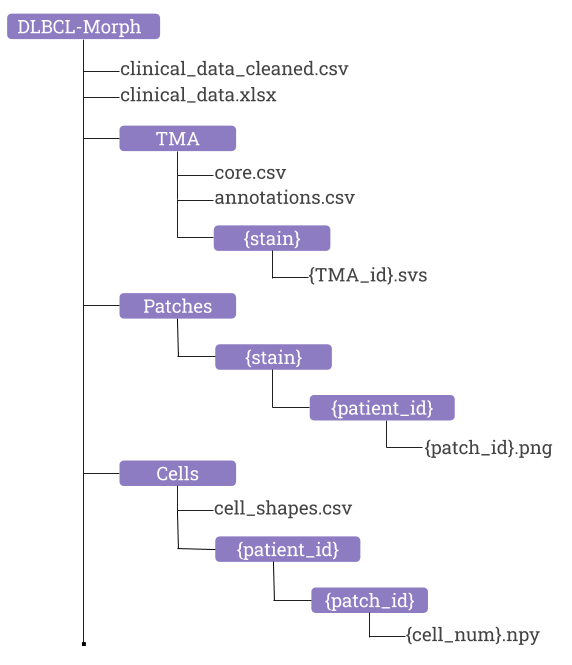}
\caption{\textbf{The directory structure of DLBCL-Morph}}
\label{fig:data_structure}
\end{figure}

The DLBCL-morph dataset is organized into 
three 
folders, 
\emph{TMA}, \emph{Patches}, and \emph{Cells} as is shown by Figure \ref{fig:data_structure}. The clinical data of the patients together with the outcome is stored in \emph{clinical\_data.xlsx} and \emph{clinical\_data\_cleaned.csv} where the latter contains all the patients for which the outcome is recorded and all categorical variables are converted to numerical values, e.g. `neg', `pos', and `no data' were converted to 0, 1, and NaN, respectively for the variable CD10 IHC. Each patient has a unique identifier. There are 209 patients recorded in \emph{clinical\_data\_cleaned.csv}. The column \emph{OS} records the overall survival which is the length of time (in years) from the end of treatment until death or last follow-up. The column \emph{Follow-up Status} (FUS) is 1 if the patient was deceased at the time of last follow-up, else 0. 



\subsection*{TMA}

The \emph{TMA} folder contains a total of 42 digitally-scanned TMAs, which are organized within subfolders for each stain. The filename of each TMA is a TMA id which is the same across all stains, i.e. \emph{DLBCL-Morph/TMA/HE/TMA255} and \emph{DLBCL-Morph/TMA/BCL2/TMA255} contains cores of the same set of patients. The TMA id together with the row and column number of each core, starting with 0 and 0, respectively in the upper left corner,  can be linked to the patient id through \emph{core.csv}, each patient has two cores. The \emph{annotations.csv} contains coordinates of ROIs annotated by human experts. For each annotation there is a patient id, TMA id, and stain where the TMA id and the stain is used to locate the TMA file that the annotation belongs to. The annotations are rectangular and the coordinates record the upper left and lower right corners based on the 40x magnification level of the TMAs.

\subsection*{Patches}

The Patches folder contains subfolders of stains which contains subfolders of patients that has at least one ROI. The patches are localized in the folders of patient ids with a patch id as the filename and are stored in PNG format. There are 195 patients that have at least one patch from at least one stained TMA. However, some patients do not contain patches for all 6 stains, which can occur if the core for a particular stain was missing or not covered by any ROIs.

\subsection*{Cells}

The \emph{Cells} folder contains subfolders of patient ids which contains subfolders of patch ids. The binary segmentation images for tumor cell nucleus are localized in the folders of patch ids with the cell number as the filename and stored in NPY format. The NPY format is used by the Numpy package for Python to save arrays, in this case we are storing 2-dimensional arrays with binary values as segmentations of tumor cell nucleus. The cell numbers are non-consecutive since all non-tumor cells are discarded in each patch. All the geometric features computed from tumor nuclei are stored in \emph{cell\_shapes.csv} and can be linked to the nucleus segmentation images through the patch id and the cell number.



%% file: Sections/technical_validation.tex
\section*{Technical Validation}
We performed survival regression using the geometric and clinical features in our dataset to measure the utility of these features in predicting prognostic outcome. This analysis was performed on the 170 patients for whom patches from H\&E stained TMAs were available. For each of the geometric features computed per tumor nucleus, we computed the mean and standard deviation across all nuclei for each patient. We then fit Cox Proportional Hazards models using the binary Follow-up Status (FUS) feature as an indicator of censoring, and the overall survival (OS) feature as the time to event or censoring (in years). We evaluated our models using Harrel's C-index \cite{10.1001/jama.1982.03320430047030}. Random prediction would give a C-index of $0.5$. Specifically we fit three models: i) using both clinical and geometric features ii) using only clinical features iii) using only geometric features.  

We used the bootstrap method to obtain an ``optimism-corrected" C-index \cite{Harrell1996}. We sampled 1000 bootstrap replicates with replacement and fit the model on each bootstrap replicate. We then evaluated the model on both the original data and the bootstrap replicate. We recorded the performance decrease between evaluating on the bootstrap replicate and evaluating on the original data. This decrease, averaged over all bootstrap replicates, was subtracted from the original C-index to obtain the optimism-corrected C-index. We also generated the corresponding 95\% two-sided confidence intervals (CI) for the optimism-corrected C-indices using the non-parametric percentile bootstrap method \cite{efron_bootstrap_1986} with 1000 bootstrap replicates.

The resulting optimism-corrected C-indices with 95\% CIs for our models were: i) $0.700$ $(0.651, 0.744)$ using clinical and geometric features, ii) $0.674$ $(0.602, 0.737)$ using only clinical features and iii) $0.635$ $(0.574, 0.691)$ using only geometric features. Thus, use of the geometric features alone allowed significantly better than random survival prediction. Use of both clinical and geometric features led to a higher performance than the use of clinical features alone, although this performance difference was not statistically significant. While prognostic classification based on the morphologic properties of the tumor has proved to be challenging and the subject of continued debate, \cite{10.1182/blood.V89.7.2291, Baars1999, Diebold2002, Nakamine1993, Salar2009, Villela2001} our results suggest that geometric features computed from H\&E-stained tumor nuclei can provide a significant signal to predict surival outcome. This finding should be further evaluated on external datasets and prospectively in future studies.


%% file: Sections/usage_notes.tex
\section*{Usage Notes}
The DLBCL-Morph dataset can be downloaded here: \href{https://stanfordmedicine.box.com/s/0sh3plpjfovea6gv93y8a5ch1k3j0lr5}{https://stanfordmedicine.box.com/s/0sh3plpjfovea6gv93y8a5ch1k3j0lr5}. The data is organized as shown in Figure \ref{fig:data_structure}. We have provided publicly available Jupyter Notebooks \cite{4160251, soton403913} to illustrate computation of geometrical features as well as usage of the data. One notebook uses the clinical and geometric variables in the dataset to reproduce the survival regression results described in the Technical Validation section. Another notebook visualizes and reproduces the computation of several geometric features for any segmented tumor nucleus in our dataset. Finally, we provide another notebook to extract patches uniformly from inside any of the ROIs in the dataset. These patches are already included as part of the dataset, but we believe this notebook will be beneficial for researchers who work with the SVS files in our dataset. The notebooks, along with the code to compute all geometrical features from tumor nuclei, are provided at \href{https://github.com/stanfordmlgroup/DLBCL-Morph}{https://github.com/stanfordmlgroup/DLBCL-Morph}.



%% file: Sections/code_availability.tex
\section*{Code availability}

The code to compute all geometric features from all tumor nuclei in our dataset, along with notebooks to illustrate usage of our data and reproduce all survival regression results, is publicly available at \href{https://github.com/stanfordmlgroup/DLBCL-Morph}{https://github.com/stanfordmlgroup/DLBCL-Morph}.
